\documentclass[final]{cvpr}

\usepackage{times}
\usepackage{epsfig}
\usepackage{graphicx}
\usepackage{amsmath}
\usepackage{amssymb}
\usepackage{amsthm}
\usepackage{booktabs}
\usepackage{multirow}
\usepackage{float}
\usepackage[numbers]{natbib}
\usepackage{subfigure}
\newtheorem{definition}{Definition}
\DeclareMathOperator*{\argmax}{arg\,max}

\usepackage{marvosym}

\usepackage[pagebackref=true,breaklinks=true,colorlinks,bookmarks=false]{hyperref}

\def\cvprPaperID{5079} % *** Enter the CVPR Paper ID here
\def\confYear{CVPR 2021}
\begin{document}

%%%%%%%%% TITLE
\title{Regressive Domain Adaptation for Unsupervised Keypoint Detection}

% \author{Junguang Jiang, Yifei Ji, Ximei Wang,Jianmin Wang, Mingsheng Long (\Letter)
% \\
% School of Software, BNRist, Tsinghua University, China \\
% Research Center for Big Data, Tsinghua University, China\\
% Beijing Key Laboratory for Industrial Big Data System and Application\\
% {\tt\small \{jjg20,jiyf17,wxm17\}@mails.tsinghua.edu.cn,\{jimwang, mingsheng\}@tsinghua.edu.cn
% }
% \and 
% Yufeng Liu \\
% Y-tech, Kuaishou Technology\\
% {\tt\small liuyufeng@kuaishou.com}
% }

\author{
    Junguang Jiang$^1$, Yifei Ji$^1$, Ximei Wang$^1$, Yufeng Liu$^2$,  Jianmin Wang$^1$, Mingsheng Long$^1$\thanks{Correspondence to: Mingsheng Long (mingsheng@tsinghua.edu.cn).} \\
    $^1$School of Software, BNRist, Tsinghua University, China  \\ 
    $^2$Y-tech, Kuaishou Technology \\ 
}
\maketitle

%%%%%%%%% ABSTRACT
\begin{abstract}
Domain adaptation (DA) aims at transferring knowledge from a labeled source domain to an unlabeled target domain. Though many DA theories and algorithms have been proposed, most of them are tailored into classification settings and may fail in regression tasks, especially in the practical keypoint detection task. To tackle this difficult but significant task, we present a method of regressive domain adaptation (RegDA) for unsupervised keypoint detection. Inspired by the latest theoretical work, we first utilize an adversarial regressor to maximize the disparity on the target domain and train a feature generator to minimize this disparity. However, due to the high dimension of the output space, this regressor fails to detect samples that deviate from the support of the source. To overcome this problem, we propose two important ideas. First, based on our observation that the probability density of the output space is sparse, we introduce a spatial probability distribution to describe this sparsity and then use it to guide the learning of the adversarial regressor. Second, to alleviate the optimization difficulty in the high-dimensional space, we innovatively convert the minimax game in the adversarial training to the minimization of two opposite goals. Extensive experiments show that our method brings large improvement by $8\%$ to $11\%$ in terms of PCK on different datasets.

\end{abstract}

%%%%%%%%% BODY TEXT
\section{Introduction}
%第一段 （为什么我们要做这个工作）
% DA的意义（略讲）
% DA+Keypoints的意义（详细讲，让读者觉得这个任务真的很重要）
% 比如标注成本高（比分类高很多）、实际需求大、标注困难（例如深度标注、遮挡情况下的标注）
Many computer vision tasks have achieved great success with the advent of deep neural networks in recent years. However, the success of deep networks relies on a large amount of labeled data \cite{imagenet}, which is often expensive and time-consuming to collect.  
Domain adaptation (DA) \cite{DA}, which aims at transferring knowledge from a labeled source domain to an unlabeled target domain, is a more economical and practical option than annotating sufficient target samples, especially in the keypoint detection tasks. 
The fast development of computer vision applications leads to huge increases in demand for keypoint detection but the annotations of this task are more complex than classification tasks, requiring much more labor work especially when the objects are partially occluded.
On the contrary, accurately labeled synthetic images can be obtained in abundance by computer graph processing at a relatively low cost~\cite{vazquez2013virtual, virtual}. Therefore, regressive domain adaptation for unsupervised keypoint detection has a promising future.

% 第二段 （有没有其他人做过相关的工作）
% 现有的DA的原理（略讲），以DANN、MCD为例
% 现有的DA方法适用于分类问题，实验发现在关键点预测问题上都失效了
% 分析原因
% Through some DA methods (such as MCD\cite{MCD}, DD\cite{MDD}), 

There are many effective DA methods for classification \cite{DAN, DANN, MCD, MDD}, but we empirically found that few methods work on regression. One possible reason is that there exist explicit task-specific boundaries between classes in classification. 
By applying domain alignment, the margins of boundaries between different classes on the target domain are enlarged, thereby helping the model generalize to the unlabeled target domain. However, the regression space is usually continuous on the contrary, \textit{i.e.}, there is no clear decision boundary. Meanwhile, although images have limited pixels, the key point is still in a \textit{large} discrete space due to a combination of different axes, posing another huge challenge for most DA methods.

\begin{figure}[!t]
\begin{center}
   \includegraphics[width=1.\linewidth]{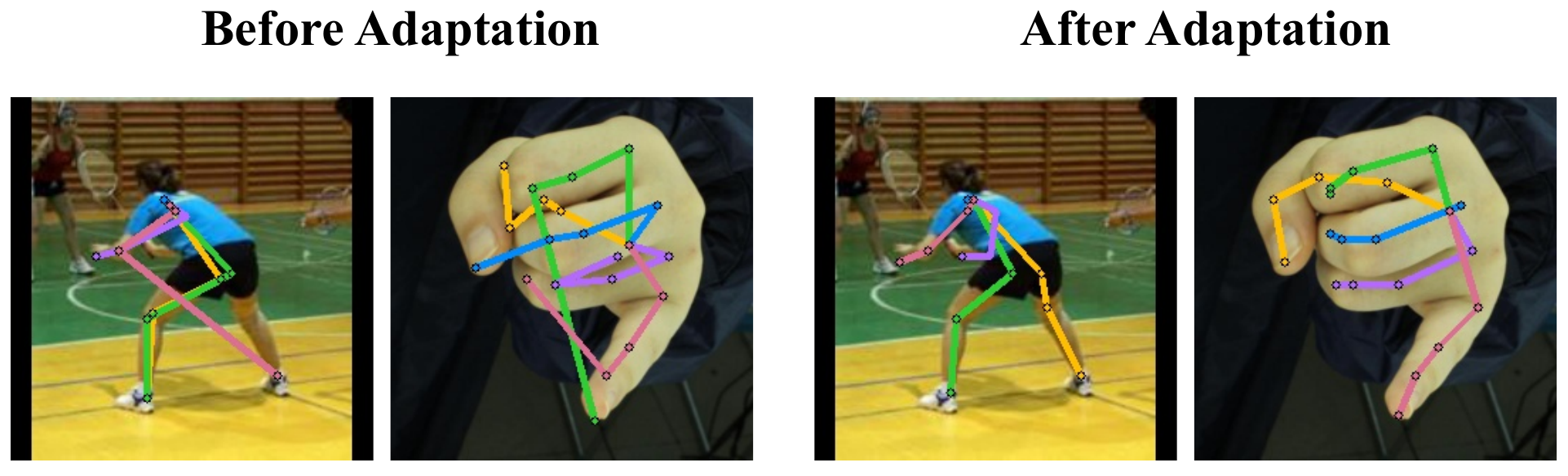}
\end{center}
\vspace{-10pt}
\caption{\textbf{Visualization} before and after adaptation on the unlabeled target domain. (Left) The wrong predictions before adaptation are usually located at other key points. (Right) The predictions of the adapted model look more like hands or bodies.}
\label{fig: introduction}
\vspace{-12pt}
\end{figure}

% 第三段 （我们的motivation是什么）
% 实验发现，在unlabeled domain上的错误主要集中在其他关键点上
To solve the issue caused by the large output space, we first delved into the prediction results of a source-only keypoint detection model. 
We unexpectedly observed that when the predictions on the unlabeled domain are wrong, they are \textit{not equally distributed} on the image.
For example, if the position of a \textit{right} ankle is mistaken (see Fig. \ref{fig: introduction}), the wrong prediction is most likely at the position of the left ankle or other key points, instead of somewhere in the background as we expected.
This observation reveals that the output space is \textit{sparse} in the sense of probability.
Consider an extremely sparse case where the predicted position is always located at a key point, then a specific ankle detection problem becomes a $K$-classification problem, and we can reduce the domain gap by enlarging the decision boundary between different key points.
This extreme case above gives us a strong hint that if we can constrain the output space 
% of the adversarial regressor 
from a whole image space into a smaller one with only $K$ key points, it may be possible to bridge the gap between regression and classification for RegDA.

% 第四段 (我们的方法是什么）
% 基于上述观察和类比，我们引入了错误概率分布，描述回归器容易犯错的位置
% 同时由于在高维空间maximize不容易帮助f'找到上述错误情况，因此我们使用两个相反目标的minimization
Inspired by the latest theoretical work of DD~\cite{MDD}, we first utilize an adversarial regressor to maximize the disparity on the target domain and train a feature generator to minimize this disparity. Based on the above observations and analyses, we introduced a spatial probability distribution to describe the sparsity and use it to guide the optimization of the adversarial regressor. It can somewhat avoid the problems caused by the large output space and reduce the gap between the keypoint detection and classification in the DA setup.
Besides, we also found that maximizing the disparity of two regressors is unbelievably difficult (see Section \ref{result: ablation}). To this end, we convert the minimax game in DD \cite{MDD} into \textit{minimization of two opposite goals}. This conversion has effectively overcame the optimization difficulty of adversarial training in RegDA. Our contributions are summarized as follows:

% 第五段 （我们的贡献）
\begin{itemize}
    \item We discovered the sparsity of the regression space in the sense of probability, which gives a hint to bridge the gap between regression and classification.
    \vspace{-5pt}
    \item 
    We proposed a new and effective method by converting the minimax game between two regressors into the minimization of two opposite goals.
    \vspace{-5pt}
    \item We conducted solid experiments on various keypoint detection tasks and prove that our method can bring performance gains by $8\%$ to $11\%$ in terms of PCK.
\end{itemize}

\section{Related Work}
\paragraph{{Domain Adaptation.}}
% DA
% DANN/DAN/MCD/MDD/CycleDA
Most deep neural networks suffer from performance degradation due to domain shift \cite{DomainShift}. Thus, domain adaptation is proposed to transfer knowledge from the source domain to the target domain.
DAN \cite{DAN} adopts adaptation layers to minimize MMD \cite{MMD} between domains. DANN \cite{DANN} first introduces adversarial training into domain adaptation. 
% CyCADA \cite{CYCADA} conducts adaption in both pixel and feature levels by imposing constraints on cycle and semantic consistency. 
MCD \cite{MCD} uses two task-specific classifiers to approximate $\mathcal{H}\Delta\mathcal{H}$-distance \cite{HHdistance} between source and target domains
and tries to minimize it by further feature adaptation. 
MDD \cite{MDD} extends the theories of domain adaptation to multiclass classification and proposes a novel measurement of domain discrepancy.
These methods mentioned above are insightful and effective in classification problems. \textit{But few of them work on regression problems.}
In our work, we propose a novel training method for domain adaptation in keypoint detection, a typical regression problem.

\vspace{5pt}
\noindent
\textbf{{Keypoint Detection.}}
% 2D Keypoints Detection
% Simple Baseline/HRNet/etc ..
%backbone
    % DeepPose: 首先使用 DNN
    % Joint training of a convolutional network and a graphical model for human pose estimation: 使用 heatmap
    % hourglass: 重复的 bottom-up, top-down 结构，对中间的 heatmap 加监督
    % CPN: 2 stage
    % HRNet: 不同分辨率的卷积模块并联；一直保留高分辨率；多次高低分辨率的特征的融合
%DA in keypoints detection
    % 1. Towards 3D Human Pose Estimation in theWild: aWeakly-supervised Approach
    % 2. 2018 ECCV Weakly-supervised 3D Hand Pose Estimation from Monocular RGB Images
    % 3. Unsupervised domain adaptation for 3d keypoint estimation via view consistency
2D keypoint detection has become a popular research topic these years for its wide use in computer vision applications.  
Tompson et al.\cite{JointTraining} propose a multi-resolution framework that generates heat maps representing per-pixel likelihood for keypoints. 
Hourglass \cite{Hourglass} develops a repeated bottom-up, top-down architecture, and enforces intermediate supervision by applying loss on intermediate heat maps. 
Xiao et al.\cite{SimpleBaselines} propose a simple and effective model that adds a few deconvolutional layers on ResNet \cite{RESNET}. HRNet \cite{HRNET} maintains high resolution through the whole network and achieves notable improvement. 
\textit{Note that our method is not intended to further refine the network architecture}, but to solve the problem of domain adaptation in 2D keypoint detection. Thus our method is compatible with any of these heatmap-based networks.

Some previous works have explored DA in keypoint detection, but most in 3D keypoints detection. Cai et al.\cite{weakly} propose a weakly-supervised method with the aid of depth images and Zhou et al.\cite{zhou2017towards} conducts weakly-supervised domain adaptation with a 3D geometric constraint-induced loss. These two methods both assume 2D ground truth on target domain available and use a\textit{ fully-supervised method} to get 2D heat map. Zhou et al. \cite{zhou2018unsupervised} utilize view-consistency to regularize predictions from unlabeled target domain in 3D keypoints detection, but depth scans and images from different views on target domain are required. \textit{Our problem setup is completely different from the above works since we only have unlabeled 2D data on the target domain.}

\vspace{5pt}
\noindent
\textbf{{Loss Functions for Heatmap Regression.}}
% Heatmap Loss相关的工作
% Integral Human Pose Estimation的第5页(In the lituratue, there are several choices of loss function for heat maps ...)列举了集中常见的Loss(竟然有Cross Entropy)和相关的论文, 我们都需要引用和对比一下
    % L2
    % cross entropy
    % binary cross-entropy
Heatmap regression is widely adopted in keypoint detection. 
Mean squared error between the predicted heat map and the ground truth is most widely used \cite{JointTraining, CPM, AdversarialPosenet, Multi-context-attention, MSPN, HRNET}.
Besides, Mask R-CNN \cite{Mask-r-cnn} adopts cross-entropy loss, where the ground truth is a one-hot heat map.
Some other works \cite{Deepercut, papandreou2017towards} take the problem as a binary classification for each pixel. 
Differently, \textit{we present a new loss function based on KL divergence}, which is suitable for RegDA.
% DA+Keypoints
% 1. Towards 3D Human Pose Estimation in theWild: aWeakly-supervised Approach
% 2. 2018 ECCV Weakly-supervised 3D Hand Pose Estimation from Monocular RGB Images
% 3.  etc. 

%-------------------------------------------------------------------------
%-------------------------------------------------------------------------
\section{Preliminaries}

\subsection{Learning Setup}
In supervised 2D keypoint detection, we have $n$ labeled samples $\{(\boldsymbol{x}_i, \boldsymbol{y}_i)\}_{i=1}^n $ from $ \mathcal{X}\times\mathcal{Y}^K$ , where $\mathcal{X} \in \mathcal{R}^{H\times W \times 3}$ is the input space, $\mathcal{Y} \in \mathcal{R}^2 $ is the output space and $K$ is the number of key points for each input.  
The samples independently drawn from the distribution $D$ are denoted as $\widehat{D}$. The goal is to find a regressor $f\in \mathcal{F}$ that has the lowest error rate $\text{err}_{D} = \mathbb{E}_{(x, y)\sim D}  L(f(x), y)$ on $D$, where $L$ is a loss function we will discuss in Section \ref{sec: method supervised keypoint detection}.

In unsupervised domain adaptation, there exists a labeled source domain $\widehat{P}=\{(\boldsymbol{x}_i^s, \boldsymbol{y}_i^s)\}_{i=1}^n$  and an unlabeled target domain $\widehat{Q}=\{\boldsymbol{x}_i^t\}_{i=1}^m$.  The objective is to minimize $\text{err}_{Q}$. 
\subsection{Disparity Discrepancy}
\begin{definition}[Disparity] 
\label{def: disparity}
Given two hypothesis $f, f^{'}\in \mathcal{F}$, we define the disparity between them as
\begin{equation}
	\mathrm{disp}_{D} (f^{'}, f)\triangleq \mathbb{E}_{D} L(f^{'}, f).
\end{equation}
\end{definition}

\begin{definition}[Disparity Discrepancy, DD] 
Given a hypothesis space $\mathcal{F}$ and a specific regressor $f \in \mathcal{F}$, the Disparity Discrepancy (DD) is defined by
\begin{equation}
\label{equ: DD definition}
\begin{split}
	d_{f, \mathcal{F}} (P, Q)&\triangleq \sup_{f^{'}\in \mathcal{F}} (\mathrm{disp}_Q(f^{'}, f)-\mathrm{disp}_{P}(f^{'},f)).\\
\end{split}
\end{equation}
\end{definition}

It has been proved that when $L$ satisfies the triangle inequality, the expected error  $\text{err}_{Q}(f)$ on the target domain is \textbf{strictly} bounded by the sum of the following four terms: empirical error on the source domain $\text{err}_{\widehat{P}}(f)$, empirical disparity discrepancy $d_{f, \mathcal{F}}(\widehat{P}, \widehat{Q})$, the ideal error $\lambda$ and complexity terms \cite{MDD}. 
Thus our task becomes 
%the following minimization problem:
\begin{equation}
	\label{equ: DD_bounds}
	\min_{f\in\mathcal{F}} \text{err}_{\widehat{P}}(f) +  d_{f, \mathcal{F}}(\widehat{P}, \widehat{Q}).
\end{equation}

\begin{figure}[htbp]
\begin{center}
   \includegraphics[width=0.7\linewidth]{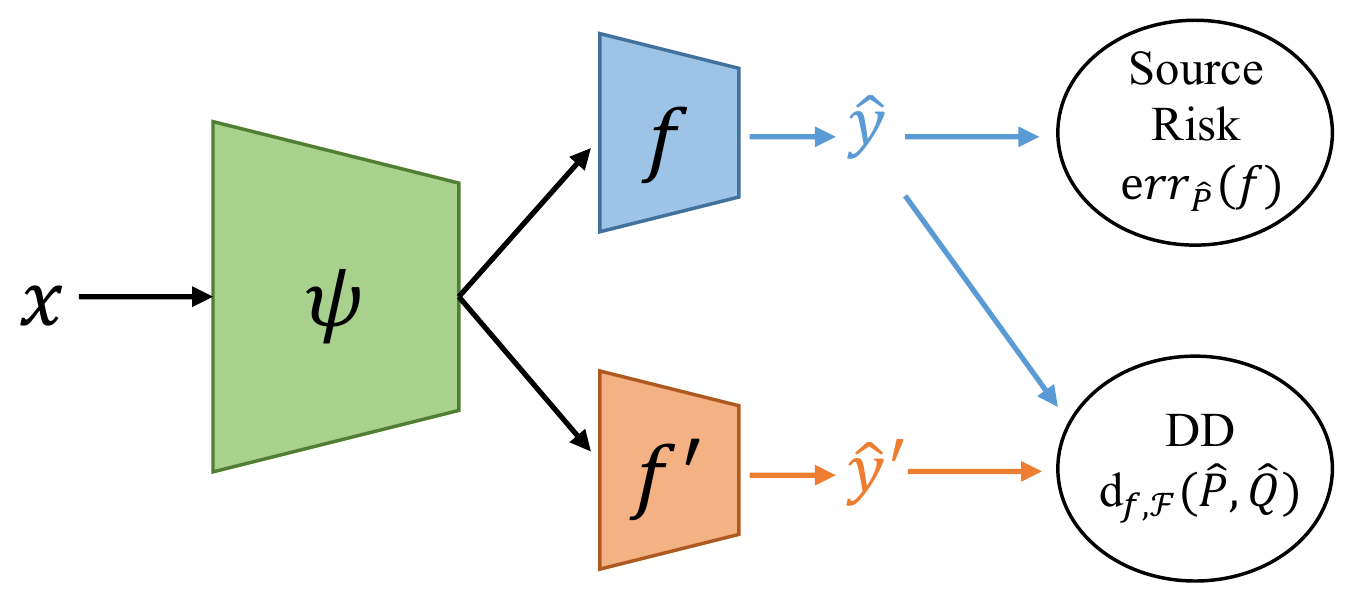}
\end{center}
\vspace{-10pt}
   \caption{DD architecture under the keypoint detection setting. 
   }
\label{fig: overall architecture}
\vspace{-5pt}
\end{figure}

We train a feature generator network $\psi$ (see Fig. \ref{fig: overall architecture}) which takes inputs $\boldsymbol{x}$, and regressor networks $f$ and $f'$ which take features from $\psi$.  
We approximate the supremum in Equation \ref{equ: DD definition} by maximizing the disparity discrepancy,
\begin{equation}
\label{Equ: DD_maximization}
\begin{split}
	\max_{f^{'}} & \mathcal{D}(\widehat{P}, \widehat{Q}) = \mathbb{E}_{\boldsymbol{x}^t\sim \widehat{Q}} L((f^{'}\circ \psi)(\boldsymbol{x}^t), (f\circ \psi)(\boldsymbol{x}^t))\\ 
    	&-\mathbb{E}_{\boldsymbol{x}^s\sim \widehat{P}} L((f^{'}\circ \psi)(\boldsymbol{x}^s), (f\circ \psi)(\boldsymbol{x}^s)).\\
\end{split}
\end{equation}
When $f^{'}$ is close to the supremum, minimizing the following terms will decrease $\text{err}_{Q}$ effectively,
\begin{equation}
\label{Equ: DD_minimization}
\begin{split}
	\min_{\psi, f} \mathbb{E}_{(\boldsymbol{x}^s, \boldsymbol{y}^s)\sim \widehat{P}} L((f\circ \psi)(\boldsymbol{x}^s), \boldsymbol{y}^s)
    + \eta \mathcal{D}(\widehat{P}, \widehat{Q}),
\end{split}
\end{equation}
where $\eta$ is the trade-off coefficient. 

\section{Method}
\subsection{Supervised Keypoint Detection}
\label{sec: method supervised keypoint detection}
Most top-performing methods on keypoint detection \cite{SimpleBaselines, HRNET, Hourglass} generate a likelihood heat map $\mathcal{H}(\boldsymbol{y}_k)\in R^{H'\times W'}$ for each key point $\boldsymbol{y}_k$ .  The heat map usually has a 2D Gaussian blob centered on the ground truth location $\boldsymbol{y}_k$. Then we can use $L_2$  distance to measure the  difference between the predicted heat map $f (\boldsymbol{x}^s)$ and the ground truth $ \mathcal{H}(\boldsymbol{y}^s)$.
The final prediction is the point with the maximum probability in the predicted map $\boldsymbol{h}_k$, i.e. $\mathcal{J}(\boldsymbol{h}_k)={\argmax_{\boldsymbol{y}\in \mathcal{Y}}}\boldsymbol{h}_k (\boldsymbol{y})$.
Heat map learning shows good performance in the supervised setting.
% L2 on Heatmap的缺点
However, when we apply it to the minimax game for domain adaptation, we empirically find that it will lead to a numerical explosion. The reason is that $f (\boldsymbol{x}^t)$ is not bounded, and the maximization will increase the value at all positions on the predicted map.

To overcome this issue, we first define the \textit{spatial probability distribution} $\mathcal{P}_{\mathrm{T}}(\boldsymbol{y}_k)$, which normalizes the heat map $\mathcal{H}(\boldsymbol{y}_k)$ over the spatial dimension,
\begin{equation}
\mathcal{P}_{\mathrm{T}}(\boldsymbol{y}_k)_{h,w} = \dfrac{\mathcal{H}(\boldsymbol{y}_k)_{h,w}}{\sum_{h'=1}^{H'} \sum_{w'=1}^{W'} \mathcal{H}(\boldsymbol{y}_k)_{h',w'}}.
\end{equation}
Denote by $\sigma$ the spatial softmax function,
\begin{equation}
\begin{split}
\sigma (\mathbf{z})_{h,w} = \dfrac{\exp{\mathbf{z}_{h,w}}}{\sum_{h'=1}^{H'}\sum_{w'=1}^{W'} \exp{\mathbf{z}_{h', w'}}} .
\end{split}
\end{equation}Then we can use KL divergence to measure the  difference between the predicted spatial probability $\widehat{\boldsymbol{p}}^s = (\sigma \circ f) (\boldsymbol{x}^s) \in R^{K\times H \times W}$ and the ground truth label $ \boldsymbol{y}^s$,
\begin{equation}
L_{\mathrm{T}} (\boldsymbol{p}^s, \boldsymbol{y}^s) \triangleq
\dfrac{1}{K} \sum_{k}^{K} \mathrm{KL}(\mathcal{P}_{\mathrm{T}}(\boldsymbol{y}_k^s) || \boldsymbol{p}^s_{k}).
\end{equation} 
In the supervised setting, models trained with KL divergence achieve comparable performance with models trained with $L_2$ since both models are provided with pixel-level supervision.  Since $\sigma(\mathbf{z})$ sums to $1$ in the spatial dimension, the maximization of $L_{\mathrm{T}} (\boldsymbol{p}^s, \boldsymbol{y}^s)$  will not cause the numerical explosion. 
In our next discussion, KL is used by default.

\subsection{Sparsity of the Spatial Density}
Compared with classification models, the output space of the keypoint detection models is much larger, usually of size $64\times 64$. Note that the optimization objective of the adversarial regressor $f'$ is to maximize the disparity between the predictions of $f'$ and $f$ on the target domain,  and minimize the disparity on the source domain.
In other words, we are looking for an adversarial regressor $f'$ which predicts correctly on the source domain,  \textit{while making as many mistakes as possible on the target domain}. 
% However, since the  the output space is very large, it would be hard to fully explore this space and find the adversarial regressor $f'$ that does poorly only on the target domain. 
However, in the experiment on \textit{dSprites} (detailed in Section \ref{sec: toy_experiments}), we find that increasing the output space of the  adversarial regressor $f'$ will worsen the final performance on the target domain. Therefore, the dimension of the output space has a huge impact on the adversarial regressor.  \textit{It would be hard to find the adversarial regressor $f'$ that does poorly only on the target domain when the output space is too large.}

\begin{figure}[htbp]
\begin{center}
   \includegraphics[width=1.\linewidth]{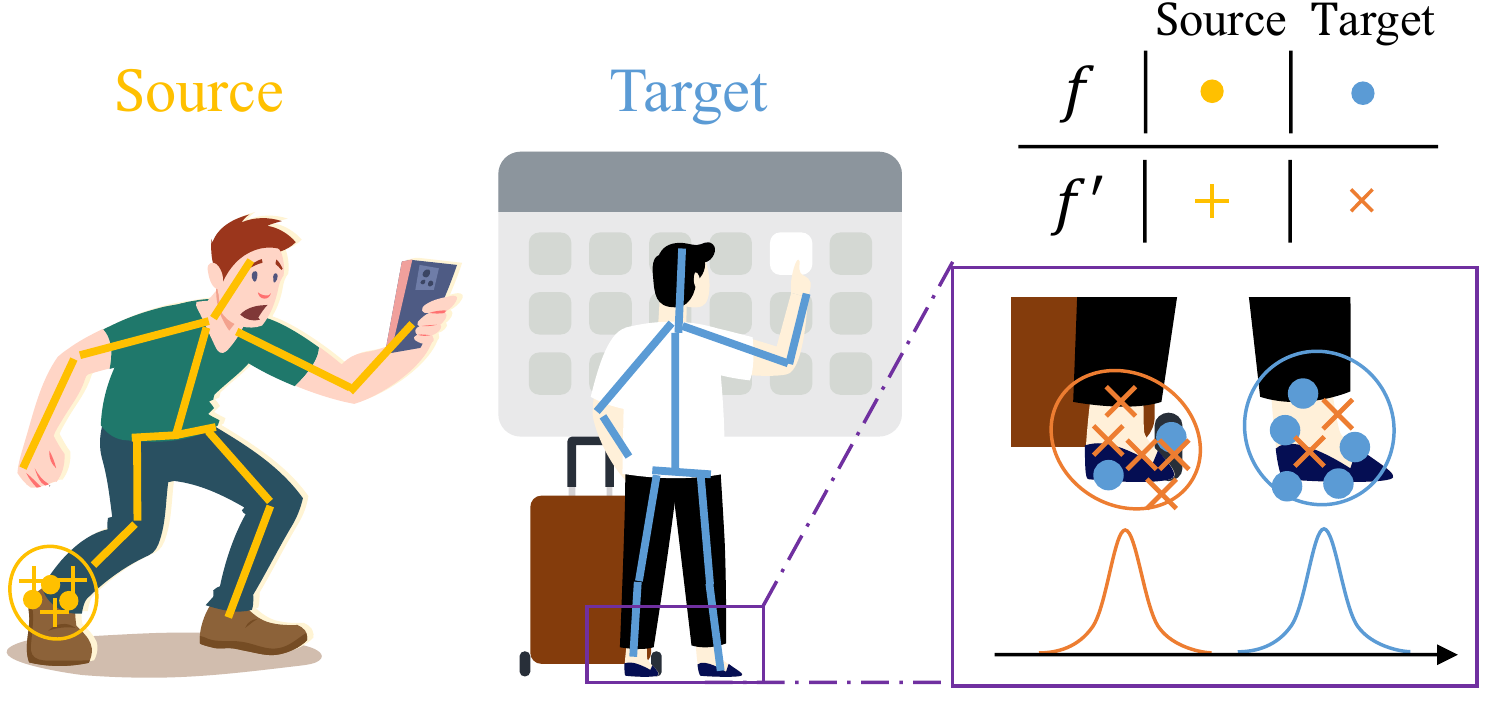}
\end{center}
\vspace{-10pt}
   \caption{The task is to predict the position of the right ankle. Predictions of $f$ and $f'$ on the source domain (in yellow) are near the right ankle. Predictions of $f$ on the target domain (in blue) are sometimes wrong and located at the left ankle or other key points. The predictions of $f'$ on the target domain (in orange) are encouraged to locate at other key points in order to detect samples far from the support of the right ankle.}
   \label{fig: ground false idea}
 \vspace{-10pt}
\end{figure}

Thus,  how to reduce the size of the output space for the adversarial regressor  has become an urgent problem.
As we mention in the introduction (see Fig.\ref{fig: introduction}), when the model makes a mistake on the unlabeled target domain, the probability of different positions is not the same. For example, when the model incorrectly predicts the position of the right ankle (see Fig. \ref{fig: ground false idea}), most likely the position of the left ankle is predicted, occasionally other key points predicted, and rarely positions on the background are predicted. 
\textit{Therefore, when the input is given, 
the output space, in the sense of probability, is not uniform. }
This spatial density is sparse, i.e. some positions have a larger probability while most positions have a probability close to zero.
% Although the adversarial regressor $f'$ cannot fully explore this large space, it can explore this space more efficiently.
To explore this space more efficiently, $f'$ should pay more attention to positions with high probability. 
Since wrong predictions are often located at \textbf{other} key points, we sum up their heat maps,  
\begin{equation}
\label{Equ: spatial probability distribution}
\mathcal{H}_{\mathrm{F}}(\widehat{\boldsymbol{y}}_k)_{h,w} = \sum_{k'\neq k} \mathcal{H}(\widehat{\boldsymbol{y}}_{k'})_{h, w},
\end{equation}
where $\widehat{\boldsymbol{y}}_k$ is the prediction by the regressor $f$.Then we normalize the map $\mathcal{H}_{\mathrm{F}}(\boldsymbol{y}_k)$,

\begin{equation}
\label{Equ: spatial probability distribution}
\mathcal{P}_{\mathrm{F}}(\widehat{\boldsymbol{y}}_k)_{h,w}  = \dfrac{\mathcal{H}_{\mathrm{F}}(\widehat{\boldsymbol{y}}_k)_{h,w}}{\sum_{h'=1}^{H'} \sum_{w'=1}^{W'} \mathcal{H}_{\mathrm{F}}(\widehat{\boldsymbol{y}}_k)_{h',w'}}.
\end{equation}
We use $\mathcal{P}_{\mathrm{F}}(\widehat{\boldsymbol{y}}_k)$  to approximate the \textit{spatial probability distribution} that the model makes mistakes at different locations and we will use it to guide the exploration of $f'$ in Section \ref{sec: minimax of target disparity}. 
The size of the output space of the adversarial regressor is reduced \textit{in the sense of expectation}.
Essentially, we are making use of the sparsity of the spatial density to help the minimax game in the high-dimensional space. 

\subsection{Minimax of Target Disparity}
\label{sec: minimax of target disparity}
Besides the problem discussed above, there is still one problem  in the minimax game of the target disparity. 
Theoretically, the minimization of KL divergence between two distributions is unambiguous. As the probability of each location in the space gets closer, two probability distribution will also get closer.
\textit{Yet the maximization of KL divergence will lead to uncertain results.} Because there are many situations where the two distributions are different, for instance, the variance is different or the mean is different.

\begin{figure}[htbp]
\begin{center}
   \includegraphics[width=0.9\linewidth]{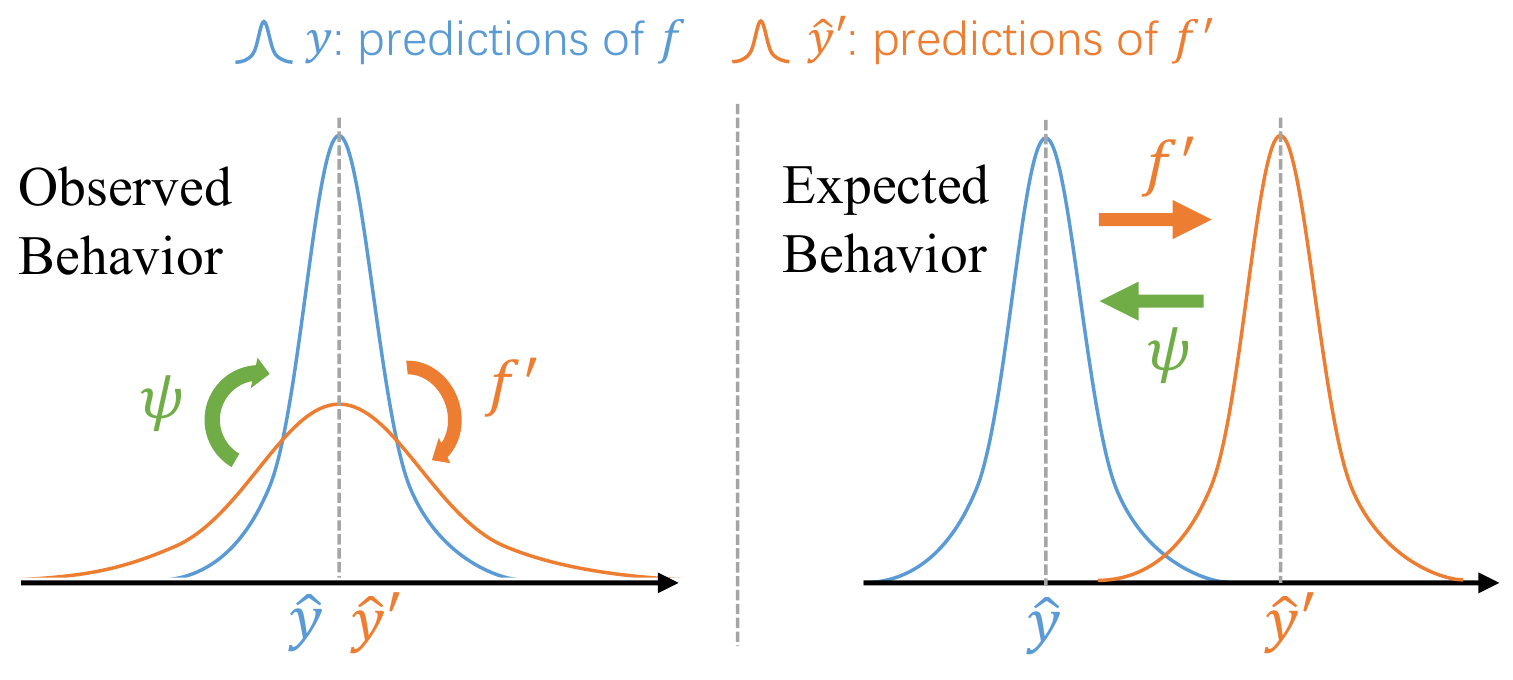}
\end{center}
\vspace{-10pt}
   \caption{When we maximize the KL between the predictions by $f'$ and $f$ (fixed), we expect to maximize the mean difference, but what actually changes is often \textit{only} the variance.}
   \label{fig: maximization of distribution}
 \vspace{-5pt}
\end{figure}

In the keypoint detection, we usually use PCK (detailed in Section \ref{pck}) to measure the quality of the model. As long as the output of the model is near the ground truth, it is regarded as a correct prediction. Therefore, we are more concerned about the target samples whose prediction is far from the true value.
In other words, we hope that after maximizing the target disparity, there is a big difference between the mean of the predicted distribution ($\widehat{\boldsymbol{y}}'$ should be different from $\widehat{\boldsymbol{y}}$ in Fig. \ref{fig: maximization of distribution}).
However, experiments show that $\widehat{\boldsymbol{y'}}$ and $\widehat{\boldsymbol{y}}$ are almost the same during the adversarial training (see Section \ref{result: ablation}). 
In other words,
\textit{maximizing KL mainly changes the variance of the output distribution. }The reason is that KL is calculated point by point in the space. When we maximize KL, the probability value of the peak point ($\widehat{\boldsymbol{y}}'$ in Fig. \ref{fig: maximization of distribution}) is reduced, and the probability of other positions will increase uniformly. Ultimately the variance of the output distribution increases, but the mean of the distribution does not change significantly, which is completely inconsistent with our expected behavior. Since the final predictions of $f'$ and $f$ are almost the same, it's hard for $f'$  to detect target samples that deviate from the support of the source. Thus, the minimax  game takes little effect. 

% 导致的实验结果
% Experiments also show that the PCK of  $f'$ and $f$ is almost the same during the adversarial training (see Section \ref{result: ablation}). In other words, the adversarial regressor $f'$ is too weak.  
% As a result, the minimax game on $L_{\mathrm{T}}$ does not lead to a significant increase in PCK on the target domain. 

Since maximizing cannot get our expected behavior, can we avoid using maximization and only use minimization in the adversarial training?
The answer is yes. 
The reason that we had to maximize before was that we only had one optimization goal. \textit{If we have two goals with opposite physical meanings, then the minimization of these two goals can play the role of minimax game.} Our task now is to design two opposite goals for the adversarial regressor and the feature generator.  
The goal of the feature generator is to minimize the target disparity or minimize the KL divergence between the predictions of $f'$ and $f$.
%$L_{\mathrm{T}}  ((\sigma \circ f^{'}\circ \psi) (\boldsymbol{x}_t), (\mathcal{J} \circ f \circ \psi) (\boldsymbol{x}_t)) $.
The objective of the adversarial regressor is to maximize the target disparity, and we achieve this by minimizing the KL divergence between the predictions of $f'$ and the \textit{ground false} predictions of $f$,
\begin{equation}
\label{Equ: L_f}
\begin{split}
L_{\mathrm{F}}  (\boldsymbol{p}', \boldsymbol{p}) \triangleq \dfrac{1}{K} \sum_{k}^{K} \mathrm{KL}(& \mathcal{P}_{\mathrm{F}} (  \mathcal{J}(\boldsymbol{p}))_k
  ||  \boldsymbol{p}'_k),\\
\end{split}
\end{equation}
where $\boldsymbol{p}'=(\sigma \circ f' \circ \psi) (\boldsymbol{x}^t) $ is the prediction of $f'$ and $\boldsymbol{p}$ is the prediction of $f$.
Compared to directly maximizing the distance from the \textit{ground truth} predictions of $f$, minimizing $L_{\mathrm{F}}$ can take advantage of the spatial sparsity and effectively change the mean of the output distribution.

% 直观理解和总结
Now we use Fig. \ref{fig: ground false idea} to illustrate the meaning of Equation \ref{Equ: L_f}. Assume we have $K$ supports for each key point in the semantic space.  The outputs on the labeled source domain (in yellow) will fall into the correct support. But for outputs on the target domain, the position of the left ankle might be regarded as the right ankle. These are the samples far from the supports. Through minimizing  $L_{\mathrm{F}}$, we mislead $f'$  to predict other key points as right ankle, which encourages the adversarial regressor $f'$ to detect target samples far from the support of the right ankle. Then we train the generator network $\psi$ to fool the adversarial regressor $f'$ by minimizing $L_{\mathrm{T}} $ on the target domain. This encourages the target features to be generated near the support of the right ankle. This adversarial
learning steps are repeated and the target features will be aligned to the supports of the source  finally.

\subsection{Overall Objectives}
\label{sec: training steps}
The final training objectives are summarized as follows. Though described in different steps, these loss functions are optimized simultaneously in a framework.

\paragraph{Objective 1}
First, we train the generator $\psi$ and regressor $f$ to detect the source samples correctly. Also, we train the adversarial regressor $f'$ to minimize its disparity with $f$ on the source domain. The objective is as follows:
\begin{equation}
\begin{split}
\min_{\psi, f, f'} &\mathbb{E}_{(\boldsymbol{x}_s, \boldsymbol{y}^s) \sim \widehat{P}}  ( L_{\mathrm{T}} ((\sigma  \circ f\circ \psi) (\boldsymbol{x}_s), \boldsymbol{y}^s)  \\
+ \eta &L_{\mathrm{T}}  ((\sigma  \circ f^{'}\circ \psi) (\boldsymbol{x}_s), (\mathcal{J} \circ f\circ \psi) (\boldsymbol{x}_s)) ).
\end{split}
\end{equation}

\begin{figure}[H]
\begin{center}
   \includegraphics[width=1\linewidth]{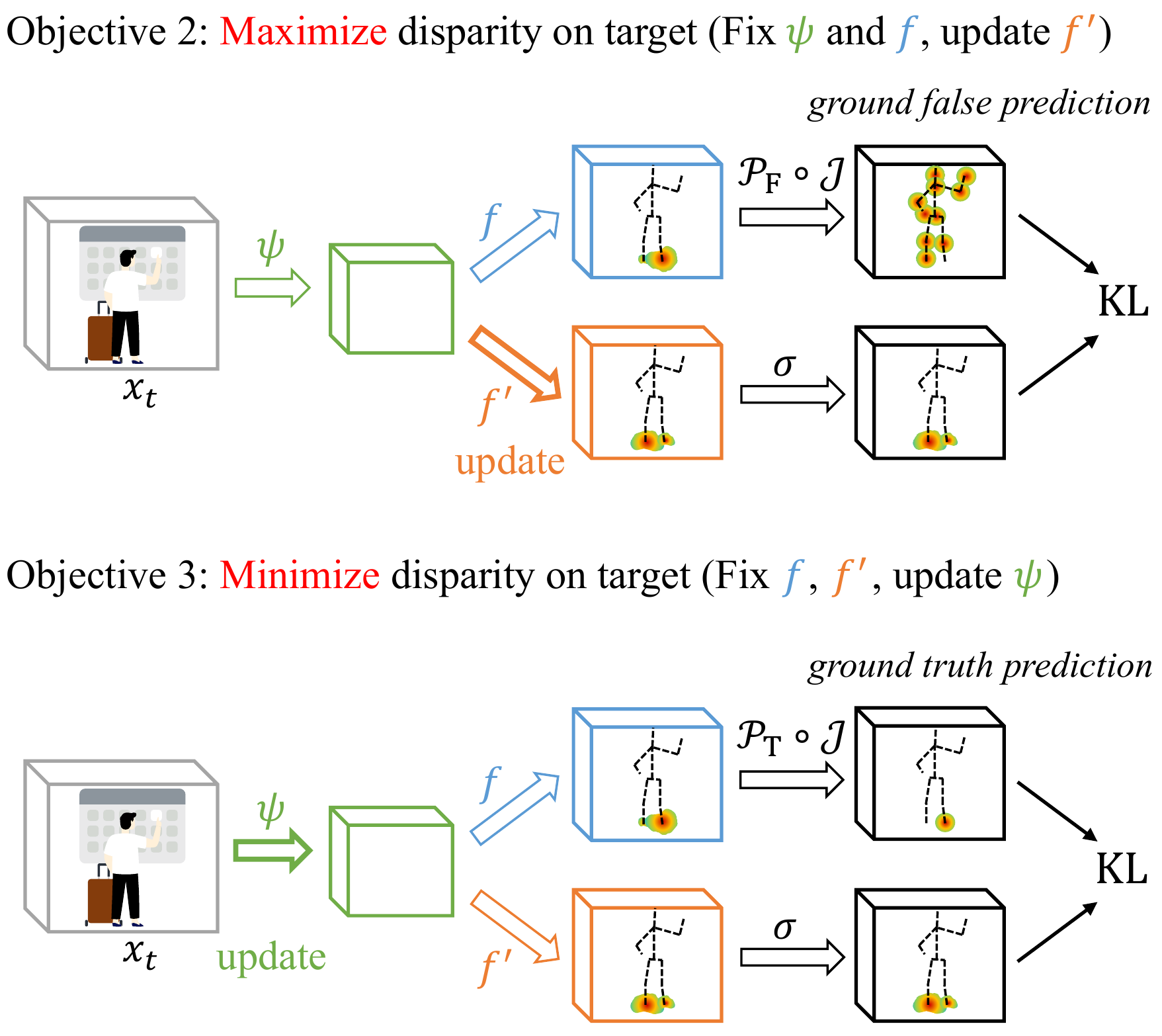}
\end{center}
 \vspace{-5pt}
 \caption{Adversarial training objectives. Our network has three parts: feature generator $\psi$, regressor $f$ and adversarial regressor $f'$ .
\textbf{Objective 2}: $f'$ learns to maximize the target disparity by minimizing its KL with \textit{ground false} predictions of $f$.  \textbf{Objective 3}: $\psi$ learns to minimize the target disparity by minimizing the KL between the predictions of $f'$ with \textit{ground truth} predictions of $f$.}
   \label{fig: training steps}
\vspace{-5pt}
\end{figure}

\paragraph{Objective 2}
Besides, we need the adversarial regressor $f'$ to increase its disparity with $f$ on the target domain by minimizing $L_{\mathrm{F}}$. By maximizing the disparity on the target domain, $f'$  can detect the target samples that deviate far from the support of the source. This corresponds to \textbf{Objective 2} in Fig. \ref{fig: training steps}, which can be formalized as follows:
\begin{equation}
	\min_{f'} \eta \mathbb{E}_{\boldsymbol{x}_t\sim \widehat{Q}} L_{\mathrm{F}} ((\sigma  \circ f^{'}\circ \psi)(\boldsymbol{x}_t), ( f\circ \psi)(\boldsymbol{x}_t)).
\end{equation}

\paragraph{Objective 3}
Finally, the generator $\psi$  needs to minimize the disparity between the fixed regressors $f$ and $f'$ on the target domain. This corresponds to \textbf{Objective 3} in Fig. \ref{fig: training steps},
\begin{equation}
	\min_{\psi}  \eta  \mathbb{E}_{\boldsymbol{x}_t\sim \widehat{Q}} \eta L_{\mathrm{T}} ((\sigma  \circ f^{'}\circ \psi)(\boldsymbol{x}_t), (\mathcal{J} \circ f\circ \psi)(\boldsymbol{x}_t)).
\end{equation}

\section{Experiments}
First, we experiment on a toy dataset called \textit{dSprites} to illustrate how dimension of the output space affects the minimax game.
Then we perform extensive experiments on real-world datasets, including hand datasets (\textit{RHD}$\rightarrow$\textit{H3D}) and human datasets (\textit{SURREAL}$\rightarrow$\textit{Human3.6M}, \textit{SURREAL}$\rightarrow$\textit{LSP}), to verify the effectiveness of our method.  We set $\eta = 1$ on all datasets.  \textbf{Code is available at \url{https://github.com/thuml/Transfer-Learning-Library/tree/dev}.}

\subsection{Experiment on Toy Datasets}
\label{sec: toy_experiments}
\begin{figure}[htbp]
\vspace{-5pt}
\begin{center}
   \includegraphics[width=0.8\linewidth]{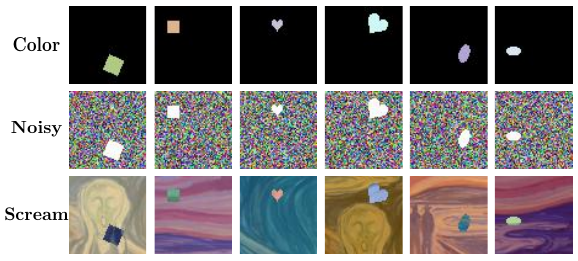}
\end{center}
\vspace{-10pt}
\caption{Some example images in the dSpirtes dataset.}
\label{fig: dSprites_dataset}
\vspace{-15pt}
\end{figure}

\paragraph{Dataset}
% 介绍数据集dSprites
\textit{DSprites} is a 2D synthetic dataset (see Fig. \ref{fig: dSprites_dataset}). It consists of three domains: \textit{Color} (C), \textit{Noisy} (N) and \textit{Scream} (S), with $737,280$ images in each. There are four regression factors and we will focus on two of them: position X and Y.  We generate a $64\times 64$ heat map for the key point. 
Experiments are performed on six transfer tasks: C→N,C→S,N→C,N→S,S→C, and S→N.

% TODO： dSprites数据集示例

\paragraph{Implementation Details} We finetune ResNet18 \cite{RESNET} pre-trained on ImageNet. Simple Baseline\cite{SimpleBaselines} is used as our detector head and is trained from scratch with learning rate $10$ times that of the lower layers. We adopt mini-batch SGD with momentum of $0.9$ and batch size of $36$. The learning rate is adjusted by $\eta_p = \eta_0 (1+\alpha p)^{-\beta}$ where $p$ is the training steps, $\eta_0=0.1, \alpha=0.0001$ and $\beta=0.75$. All models are trained for $20k$ iterations and we only report their \textit{final} MAE on the target domain.
We compare our method mainly with \textbf{DD} \cite{MDD}, which is designed for classification.  We extend it to keypoint detection by replacing cross entropy loss with $L_{\mathrm{T}} $. The main regressor $f$ and the adversarial regressor $f'$ in \textbf{DD} and \textbf{ our method} are both $2$-layer convolution neural networks with width $256$. 

\paragraph{Discussions}
% 强调我们的方法是两个min, 
% The key difference between our method and DD is that DD performs minimax on the same goal, while our method designs two opposite losses and then minimize them separately. 
Since each image has only one key point in \textit{dSprites}, we cannot generate $\mathcal{P}_{\mathrm{F}}(\widehat{\boldsymbol{y}})$ according to Equation \ref{Equ: spatial probability distribution}. 
Yet we find that for each image in the \textit{dSprites}, key points only appear in the middle area $A=\{(h,w)|16\leq h \leq 47,  16\leq w \leq 47\}$.  Therefore, we only assign  positions inside $A$ with positive probability,
\begin{equation}
\label{Equ: spatial probability distribution2}
\begin{split}
\mathcal{H}_{\mathrm{F}}(\widehat{\boldsymbol{y}})_{h,w} &= \sum_{\boldsymbol{a}\in A, \boldsymbol{a} \neq \widehat{\boldsymbol{y}} } \mathcal{H}(\boldsymbol{a})_{h, w} \\ 
\mathcal{P}_{\mathrm{F}}(\widehat{\boldsymbol{y}})_{h,w} & = \dfrac{\mathcal{H}_{\mathrm{F}}(\widehat{\boldsymbol{y}})_{h,w}}{\sum_{h'=1}^{H'} \sum_{w'=1}^{W'} \mathcal{H}_{\mathrm{F}}(\widehat{\boldsymbol{y}})_{h',w'}}.
\end{split}
\end{equation}
% 因此我们的方法只是将MDD的对抗空间从64x64变成了32x32
% 但是在dSprites上误差却又显著的下降!
We then minimize $L_{\mathrm{F}} $ to maximize the target disparity. Note that Equation \ref{Equ: spatial probability distribution2} just narrows the origin  space from $64\times 64$ to $32\times 32$. 
However, this conversion from maximization to minimization has achieved significant performance gains on \textit{dSprites}. Table \ref{Results: dSprites} shows that this conversion reduces the error by $\textbf{63}\%$ in a relative sense. 
We can conclude several things from this experiment:
\begin{enumerate}
\item The dimension of the output space has a huge impact on the minimax game of the adversarial regressor $f'$. As the output space enlarges, the maximization of $f'$ would be increasingly difficult.
\vspace{-5pt}
\item When the probability distribution of the output by $f'$ is not uniform and our objective is to maximize the disparity on the target domain, minimizing the distance with this \textit{ground false} distribution is more effective than maximizing the distance with the \textit{ground truth}.
\end{enumerate}

% 实验结果包括Source Only/Ours
\begin{table}[htbp]
\caption{MAE on \textit{dSprites} for different source and target domains (lower is better). The last row (oracle) corresponds to training on the target domain with supervised data (lower bound). }
\vspace{-5pt}
\addtolength{\tabcolsep}{-3pt}
\begin{small}
\begin{center}
\begin{tabular}{@{}l|cccccc|l@{}}
\toprule
Method  & C$\rightarrow$N  &C$\rightarrow$S   &N$\rightarrow$C   & N$\rightarrow$S  &  S$\rightarrow$C & S$\rightarrow$N  & Avg \\ \midrule
ResNet18 \cite{SimpleBaselines} & 0.495 &	0.256	& 0.371 &	0.639	& 0.030	&0.090	&0.314    \\
DD \cite{MDD}  &   0.037 &	0.078	&0.054&	0.239&	0.020	& 0.044	&0.079    \\
\textbf{RegDA (ours)}       &\textbf{0.020} &	\textbf{0.028}& \textbf{0.019}&	\textbf{0.069}&	\textbf{0.014}&\textbf{0.022}	&  \textbf{0.029}   \\
\hline
Oracle  &    0.016 &	0.022 &	0.014&	0.022&	0.014&	0.016	&0.017     \\ \bottomrule
\end{tabular}
\end{center}
\end{small}
\label{Results: dSprites}
\vspace{-10pt}
\end{table}

\subsection{Experiment on Hand Keypoint Detection}
\subsubsection{Dataset}
\begin{figure}[htbp]
\vspace{-5pt}
\begin{center}
   \includegraphics[width=0.9\linewidth]{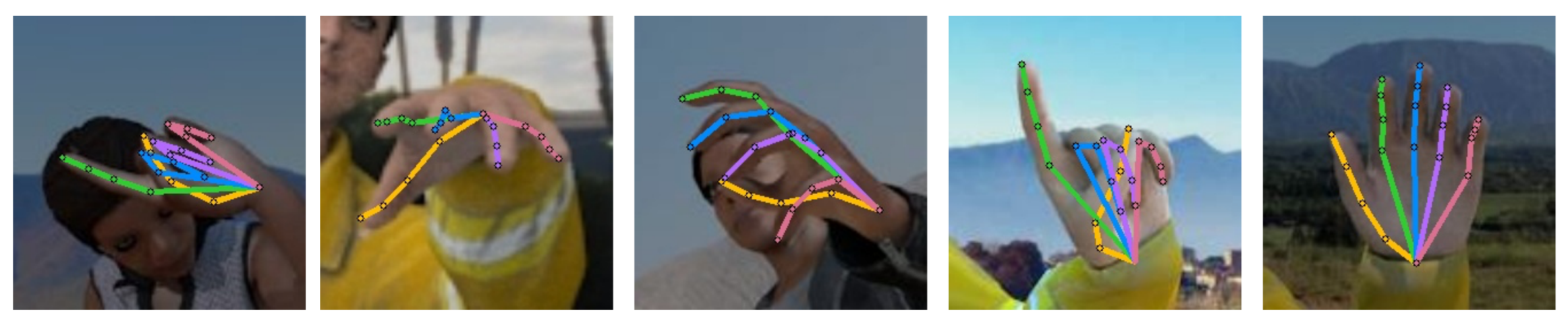}
\end{center}
\vspace{-5pt}
\caption{Some annotated images in the \textit{RHD} dataset.}
\label{fig: rhd_dataset}
\vspace{-10pt}
\end{figure}

\paragraph{RHD}
Rendered Hand Pose Dataset (\textit{RHD}) \cite{RHD} is a synthetic dataset containing $41,258$ training images and $2,728$ testing images, which provides precise annotations for $21$ hand keypoints. It covers a variety of viewpoints and difficult hand poses, yet hands in this dataset have very different appearances from those in reality (see Fig. \ref{fig: rhd_dataset}).

\paragraph{H3D}
Hand-3D-Studio (\textit{H3D}) \cite{Hand-3D-Studio} is a real-world dataset containing hand color images with $10$ persons of different genders and skin colors, $22k$ frames in total.
We randomly pick $3.2k$ frames as the testing set, and the remaining part is used as the training set. 
Since the images in \textit{H3D} are sampled from videos, many images share high similarity in appearance. Thus models trained on the training set of \textit{H3D} (\textbf{oracle}) achieve high accuracy on the testing set.  This sampling strategy is reasonable in the DA setup since we cannot access the label on the target domain.

% \paragraph{Data Preprocessing}
% % 介绍数据集处理
% Since our task is keypoint detection instead of object detection, we crop the images such that the hand is at the center of the image. The cropped images will be 1.5 times larger than the hand. Data augmentation includes scale($\pm$40\%), rotation($\pm$180 degrees) and random crop. Also ,we will generate a $64\times 64$ heatmap for each labeled key points $y_s$. 

\subsubsection{Training Details}
\label{sec: hand_training_details}
% backbone + 训练过程
We evaluate the performance of Simple Baseline \cite{SimpleBaselines} with ResNet101 \cite{RESNET} as the backbone.
% The ResNet backbone is initialized by pre-training on ImageNet. 

\textbf{Source only} model is trained with $L_2$. All parameters are the optimal parameters under the supervised setup. The base learning rate is 1e-3. It drops to 1e-4 at $45$ epochs and 1e-5 at $60$ epochs. There are $70$ epochs in total. Mini-batch size is $32$. There are $500$ steps for each epoch. Note that 
$70$ epochs are completely enough for the models to converge both on the source and the target domain. Adam \cite{Adam} optimizer is used (we find that SGD \cite{SGD} optimizer will reach a very low accuracy when combined with $L_2$).

In our method, Simple Baseline is first trained with $L_T$, with the same learning rate scheduling as \textbf{source only}. Then the model is adopted as the feature generator $\psi$ and trained with the proposed minimax game for another $30$ epochs. The main regressor $f$ and the adversarial regressor $f'$ are both $2$-layer convolution neural networks with width $256$. The learning rate of the regressor is set $10$ times to that of the feature generator, according to \cite{DANN}. For optimization, we use the mini-batch SGD with the Nesterov momentum $0.9$. 

We compare our method with several feature-level DA methods, including \textbf{DAN} \cite{DAN}, \textbf{DANN} \cite{DANN}, \textbf{MCD} \cite{MCD} and \textbf{DD} \cite{MDD}.  All methods are trained on the source domain for $70$ epochs
and then finetunes with the unlabeled data on the target domain for $30$ epochs. We report the \textit{final} PCK of all methods for a fair comparison.

\subsubsection{Results}
% 评价指标
\label{pck}
Percentage of Correct Keypoints (PCK) is used for evaluation. An estimation is considered correct if its distance from the ground truth is less than a fraction $\alpha=0.05$ of the image size. We report the average PCK on all $21$ key points. We also report PCK at different parts of the hand, such as metacarpophalangeal (MCP), proximal interphalangeal (PIP), distal interphalangeal (DIP), and fingertip.

The results are presented in Table \ref{result: rhd2h3d}. 
In our experiments, most of the existing DA methods do poorly on the practical keypoint detection task. They achieve a lower accuracy than \textbf{source only}, and their accuracy on the test set varies greatly during the training.  
In comparison, our method has significantly improved the accuracy at all positions of hands, and the average accuracy has increased by $\textbf{10.7}\%$. 

% 实验结果包括RHD->H3D
% 实验结果罗列MCP/PIP/DIP/fingertip/wrist/all, 比较ours/source only/DAN/DANN
\begin{table}[htbp]
\caption{PCK on task \textit{RHD}→\textit{H3D}.  The last row (oracle) corresponds to training on \textit{H3D} with supervised data (upper bound on the DA performance). For all kinds of key points, our approach outperforms \textbf{source only} considerably.}
\vspace{-5pt}
\begin{center}
\begin{tabular}{@{}l|cccc|l@{}}
\toprule
Method           & MCP & PIP & DIP & Fingertip & Avg \\ \midrule
ResNet101 \cite{SimpleBaselines} &   67.4  &  64.2   &  63.3   & 54.8         & 61.8     \\
DAN \cite{DAN}        &   59.0  & 57.0    &  56.3   &   48.4        &  55.1    \\
DANN \cite{DANN}       &  67.3   &  62.6   &  60.9   &   51.2        &  60.6    \\
MCD \cite{MCD}        &  59.1   &  56.1   & 54.7    &  46.9         &  54.6    \\
DD \cite{MDD}       & 72.7    &  69.6   &  66.2   &    54.4       &   65.2   \\
\textbf{RegDA (ours)}        &  \textbf{79.6}   &   \textbf{74.4}  &\textbf{71.2}    &  \textbf{62.9}          &   \textbf{72.5}   \\
\hline
Oracle      &  97.7   &   97.2  &   95.7  &    92.5      &   95.8   \\ \bottomrule
\end{tabular}
\label{result: rhd2h3d}
\end{center}
\vspace{-5pt}
\end{table}

\begin{figure}[htbp]
\begin{center}
   \includegraphics[width=1.\linewidth]{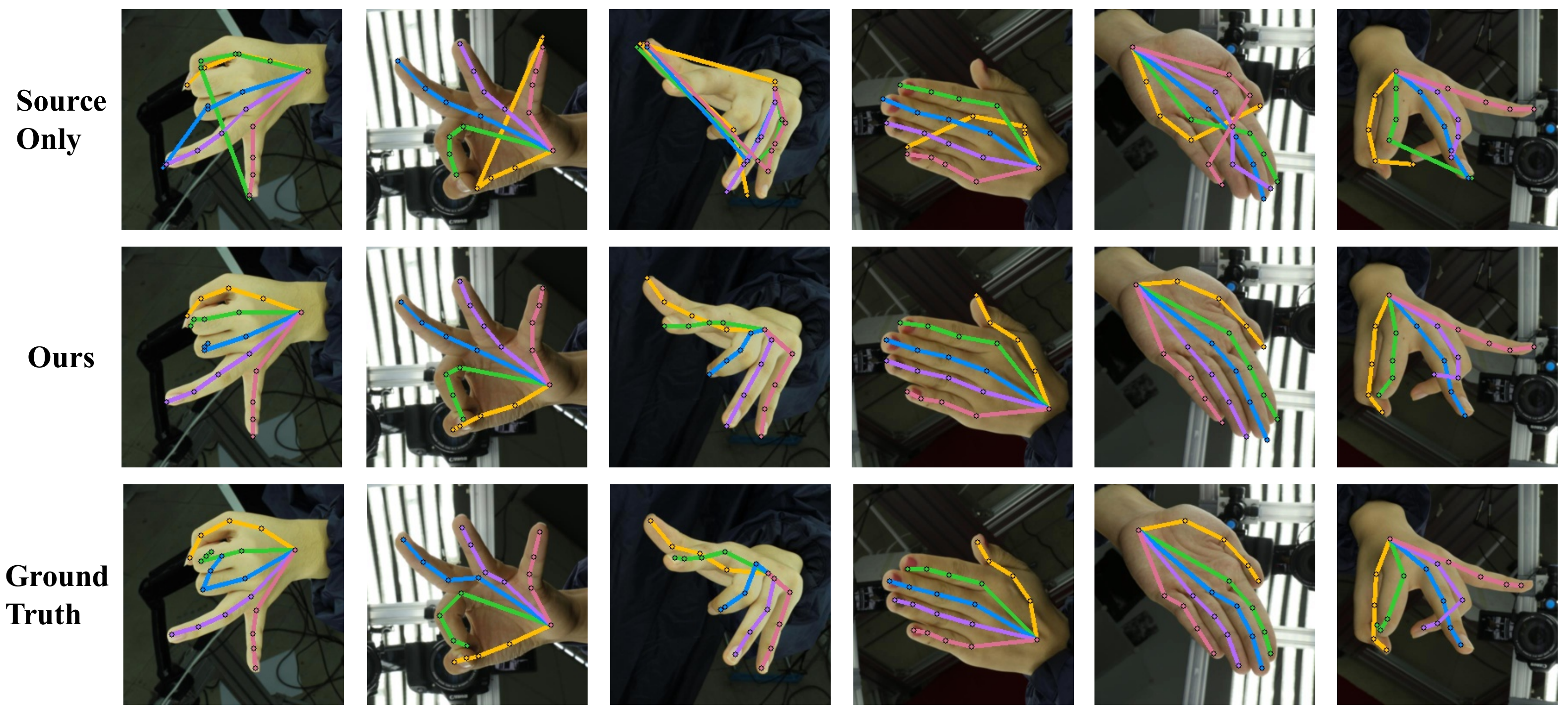}
\end{center}
\vspace{-10pt}
\caption{Qualitative results of some images in the \textit{H3D} dataset.}
\label{fig: hand_vis}
\vspace{0pt}
\end{figure}

% 可视化adaptation前后的结果
We visualize the results before and after adaptation in Fig \ref{fig: hand_vis}. 
As we mention in the introduction, the false predictions of \textbf{source only} are usually located at the positions of other key points, resulting in the predicted skeleton not look like a human hand.  
To our surprise, although we did not impose a constraint (such as bone loss \cite{zhou2017towards}) on the output of the model, the outputs of the adapted model look more like a human hand automatically.  

% 我们也对minimax的方式进行了ablation study，它们的区别见表格
% 实验结果说明，使用两个mininimize是最有效的。最大化某一项的时候会导致一项变弱。
\subsubsection{Ablation Study}
\label{result: ablation}
We also conduct an ablation study to illustrate how minimization and maximization influences adaptation.  Table \ref{result: rhd2h3d_ablation} shows the results. The first row is \textbf{DD}, which plays the minimax game on $L_{\mathrm{T}}$. The second row plays the minimax game on $L_{\mathrm{F}}$. The last row is our method, which minimizes two opposite goals separately. 
Our proposed method outperforms the previous two methods by a large margin.
\begin{table}[htbp]
\caption{Ablation study on the minimax of target disparity. }
\vspace{-5pt}
\addtolength{\tabcolsep}{-4.2pt}
\begin{center}
\begin{tabular}{@{}l|ll|cccc|l@{}}
\toprule
Method     & $f'$ & $\psi$     & MCP & PIP & DIP & Fingertip & Avg \\ \midrule
DD \cite{MDD}   & max $L_{\mathrm{T}}$ &  min $L_{\mathrm{T}}$  & 72.7    &  69.6   &  66.2   &    54.4       &   65.2   \\
& min $L_{\mathrm{F}}$&  max $L_{\mathrm{F}}$   &  74.4   &  71.1   &   66.9  &     56.4      &   66.5   \\
\textbf{RegDA}    &min $L_{\mathrm{F}}$ & min $L_{\mathrm{T}}$  &  \textbf{79.6}   &   \textbf{74.4}  &\textbf{71.2}    &  \textbf{62.9}          &   \textbf{72.5}  \\
\hline
\end{tabular}
\label{result: rhd2h3d_ablation}
\end{center}
\vspace{-10pt}
\end{table}

\begin{figure}[htbp]
\centering
\subfigure[Accuracy of  $f$]{
\includegraphics[width=0.47\columnwidth]{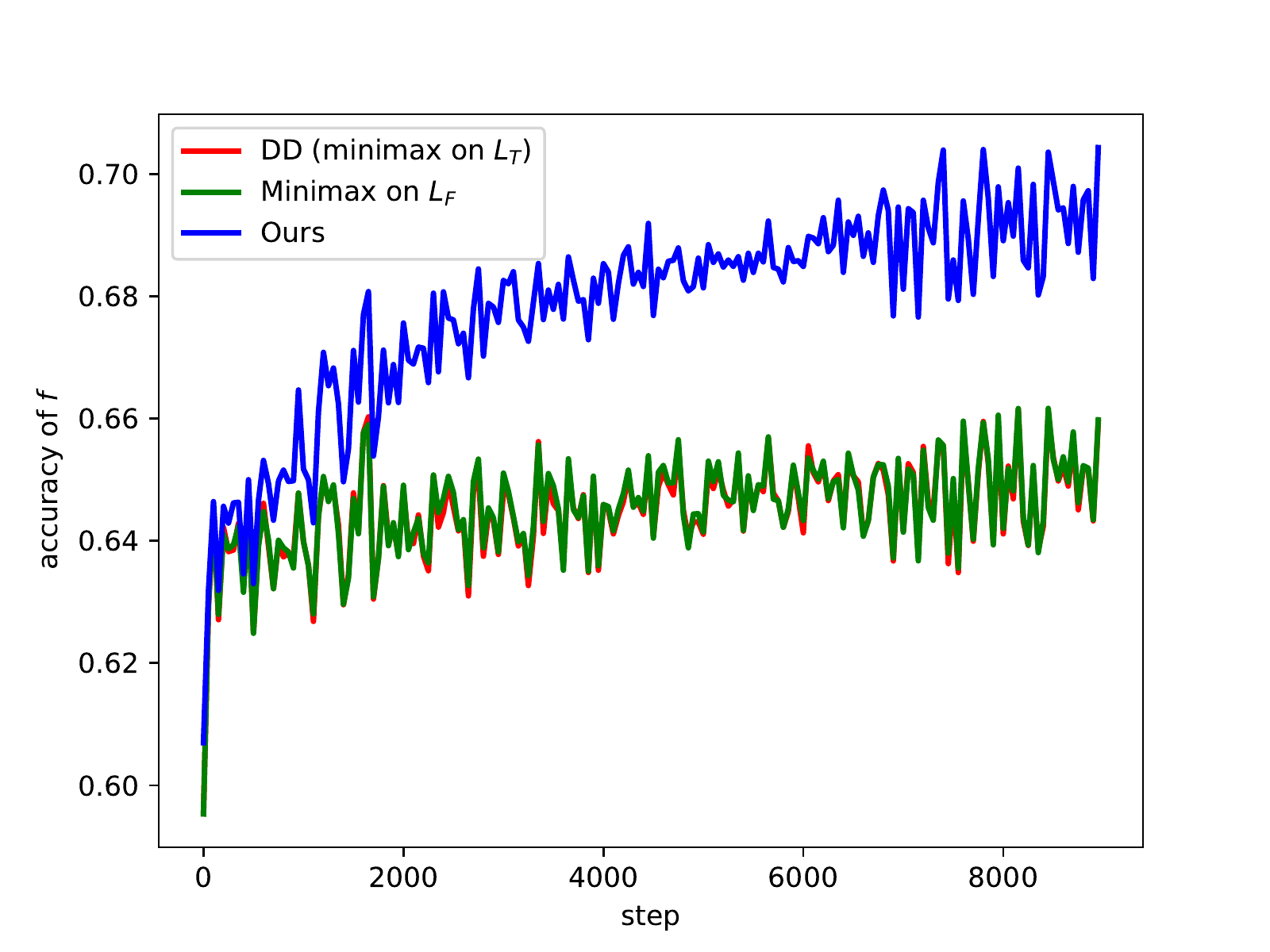}
\vspace{-10pt}
}
\subfigure[Accuracy of  $f'$]{
\includegraphics[width=0.47\columnwidth]{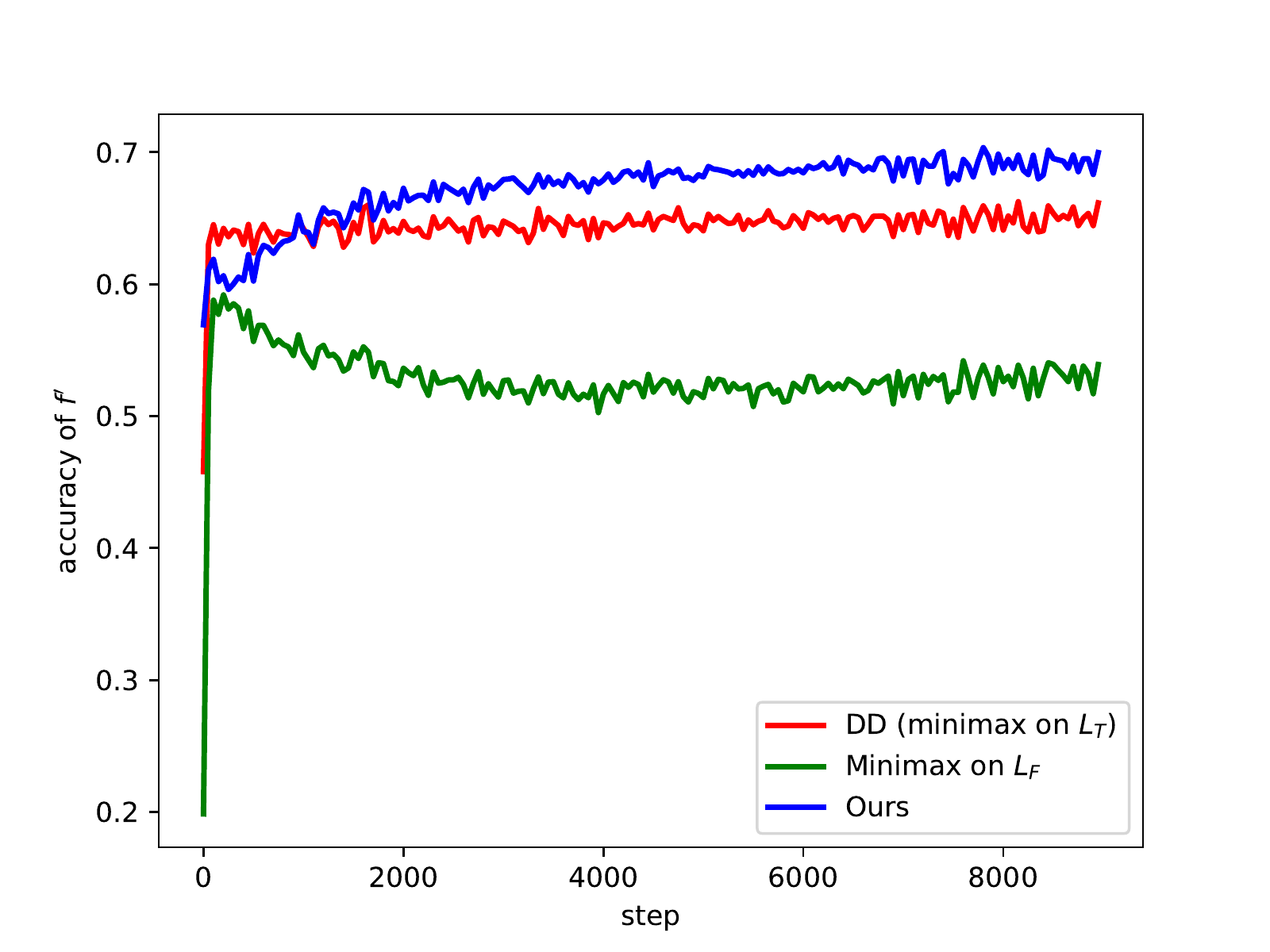}
\vspace{-10pt}
}
\subfigure[Accuracy difference]{
\includegraphics[width=0.47\columnwidth]{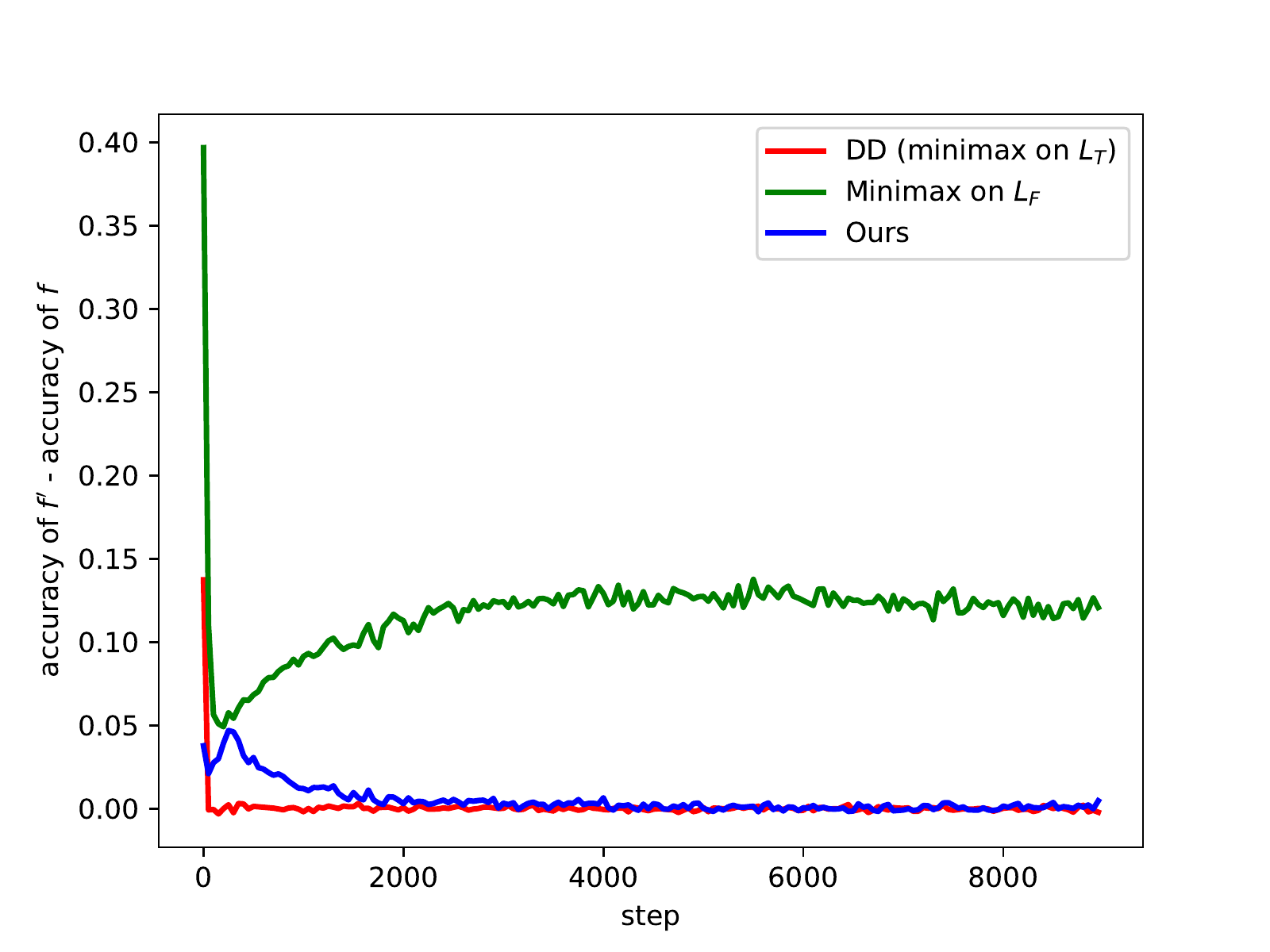}
}
\subfigure[Prediction difference]{
\includegraphics[width=0.47\columnwidth]{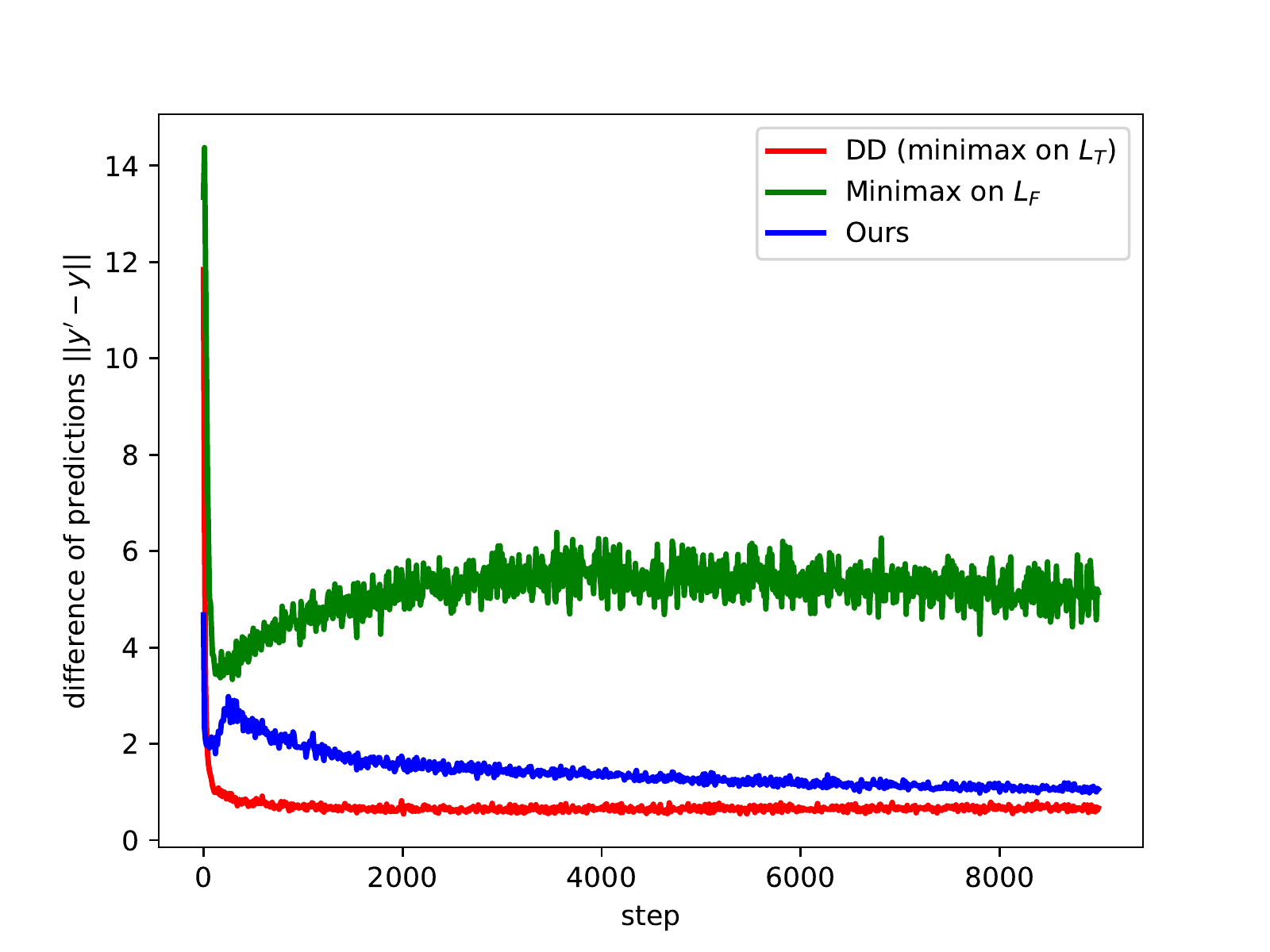}
}
\caption{Empirical values during the training process.}
\vspace{-5pt}
\label{fig:ablation}
\end{figure}

Fig. \ref{fig:ablation} visualizes the training process. For \textbf{DD}, the difference in predictions ($||\widehat{\boldsymbol{y}}'-\widehat{\boldsymbol{y}}||$) is small throughout the training process, which means that maximizing  $L_{\mathrm{T}}$ will make the adversarial regressor $f'$ weak. For methods that play minimax on $L_{\mathrm{F}}$ ,the difference in predictions keeps enlarging and the performance of $f'$ gradually drops. Thus, maximizing  $L_{\mathrm{F}}$  will make the generator $\psi$ too weak. In contrast, the prediction difference of our method increases at first and then gradually converges to zero during the adversarial training. 
As the training progresses, the accuracy of both $f$ and $f'$ \textbf{steadily} increases on the target domain.
Therefore, using two minimizations is the most effective way to do adversarial training in a large discrete space.

\subsection{Experiment on Human Keypoint Detection}
We further evaluate our method on the human keypoint detection task. The training details are the same as \ref{sec: hand_training_details}.
% and we will omit this part in this section.
\vspace{-5pt}
\subsubsection{Dataset}
\begin{figure}[htbp]
\vspace{-5pt}
\begin{center}
   \includegraphics[width=0.9\linewidth]{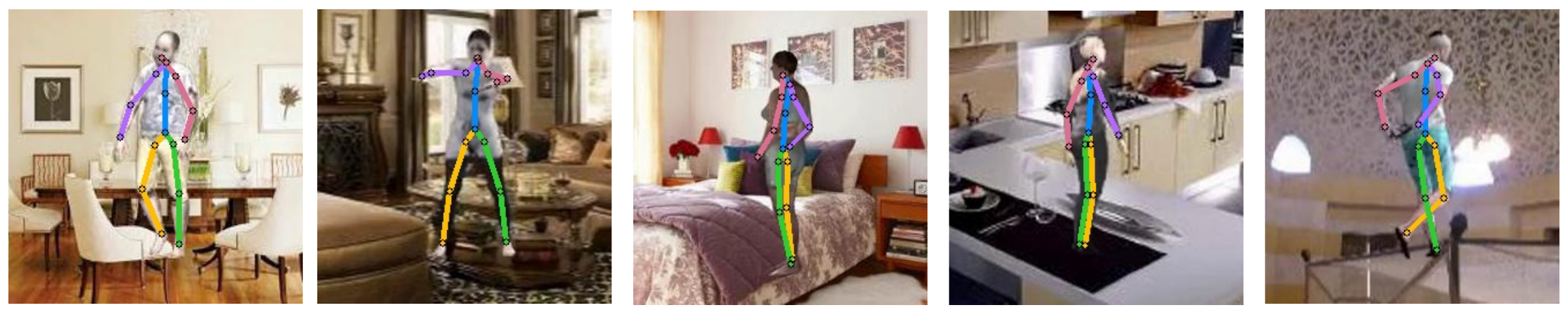}
\end{center}
\vspace{-10pt}
\caption{Some annotated images in the \textit{SURREAL} dataset.}
\label{fig: surreal_dataset}
\vspace{-10pt}
\end{figure}

\paragraph{SURREAL}
% TODO: 介绍数据集 SURREAL
\textit{SURREAL} \cite{SURREAL} is a synthetic dataset that consists of monocular videos of people in motion against indoor backgrounds (see Fig. \ref{fig: surreal_dataset}). 
There are more than 6 million frames in \textit{SURREAL}.
% the synthetic human characters are animated using Human3.6M \cite{Human3.6M}.

\paragraph{Human3.6M}
% TODO: 介绍数据集 Human3.6M，强调数据量大
\textit{Human3.6M} \cite{Human3.6M} is a large-scale real-world video dataset captured in indoor environments, with $3.6$ million frames in total.
It contains videos of human characters performing actions. We down-sampled the video from $50$\textit{fps} to $10$\textit{fps} to reduce redundancy. Following the standard in \cite{Li20143DHP}, we use 5 subjects (S1, S5, S6, S7, S8) for training and the rest $2$ subjects (S9, S11) for testing. 
\vspace{-10pt}
\paragraph{LSP}
% TODO: 介绍数据集 LSP，强调domain gap大，LSP是真实、野外数据集，而SURREAL则是虚拟的、室内数据集
Leeds Sports Pose (\textit{LSP}) \cite{LSP} is a real-world dataset containing $2k$ images with annotated human body joint locations collected from sports activities. The images in \textit{LSP} are captured in the wild, which look very different from those indoor synthetic images in \textit{SURREAL}. 
% We randomly split the dataset and use the first $80\%$ for training and the remaining $20\%$ as the test set. For a fair comparison, we keep split the same for different methods.

% \paragraph{Data Preprocessing}
% We crop the images such that the body is at the center of the image. The cropped images will be 1.5 times larger than the body. Data augmentation includes scale($\pm$25\%), rotation($\pm$30 degrees) and random crop. We generate a $64\times 64$ heatmap for each labeled key points $y_s$. 

\subsubsection{Results}
For evaluation, we also use the PCK defined in \ref{pck}. Since the key points defined by different datasets are different, we select the shared key points (such as shoulder, elbow, wrist, hip, knee) and report their PCK.

As shown in Table \ref{result: surreal2human}  and \ref{result: surreal2lsp},  our proposed method substantially outperforms \textbf{source only} at all positions of the body.
The average accuracy has increased by $\textbf{8.3\%}$ and $\textbf{10.7\%}$ on \textit{Human3.6M} and \textit{LSP} respectively.  

\begin{figure}[htbp]
\begin{center}
   \includegraphics[width=1.\linewidth]{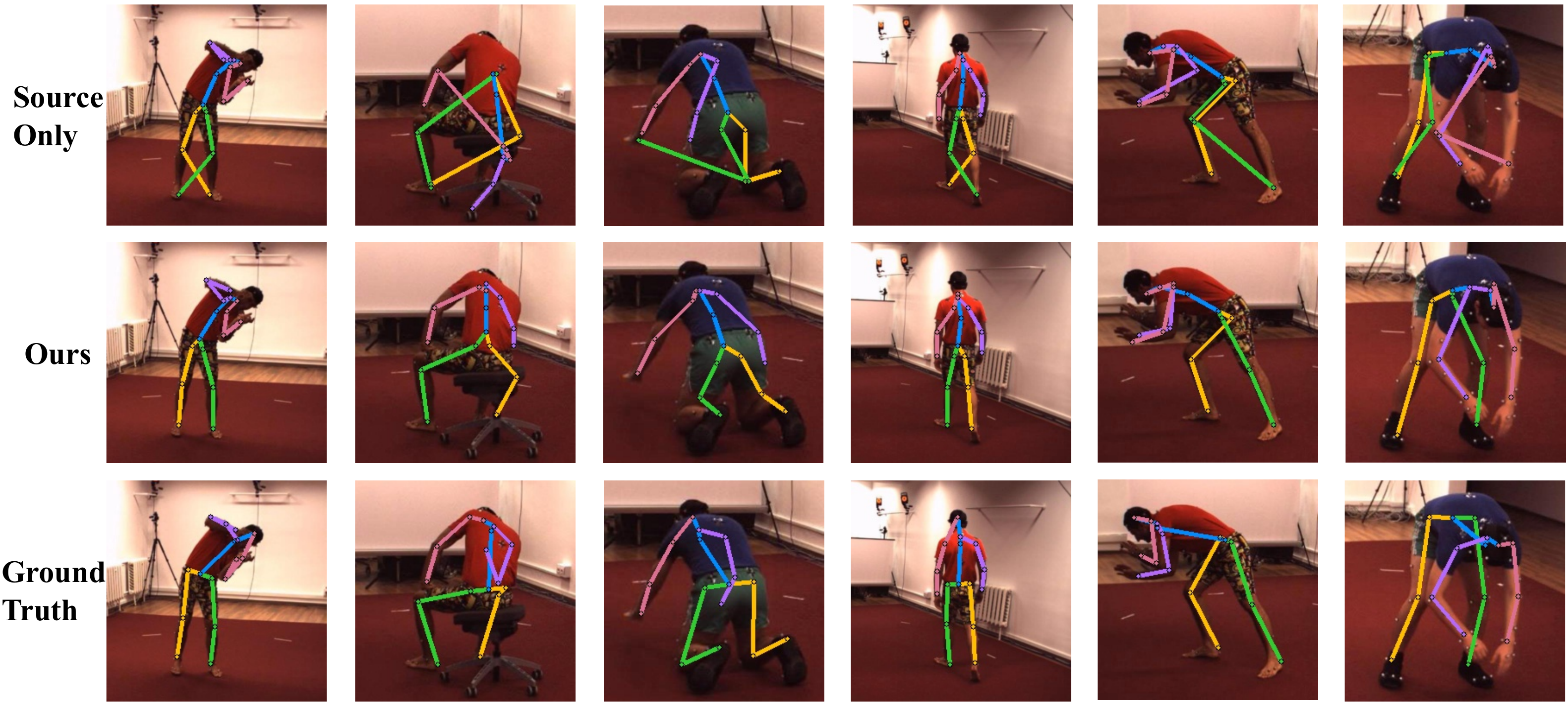}
\end{center}
\vspace{-10pt}
\caption{Qualitative results of some images in the \textit{Human3.6M} dataset. 
Note that the key points on the blue lines are not shared between different datasets.}
\label{fig: human_vis}
\end{figure}

\begin{figure}[htbp]
\begin{center}
   \includegraphics[width=1.\linewidth]{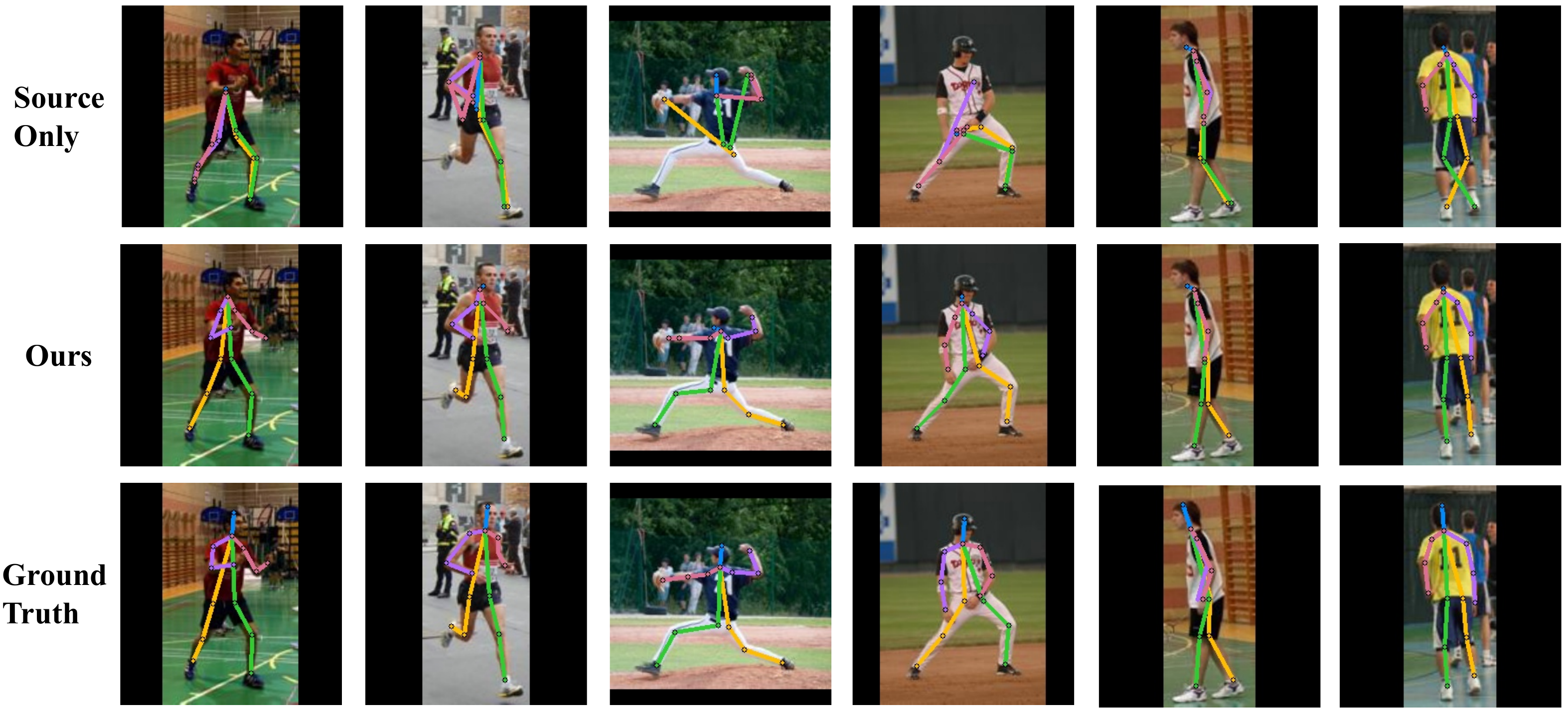}
\end{center}
\vspace{-10pt}
\caption{Qualitative results of some images in the \textit{LSP} dataset. Note that the key points on the blue lines are not shared between different datasets.}
\label{fig: lsp_vis}
\end{figure}

Fig. \ref{fig: human_vis} and  \ref{fig: lsp_vis} show the visualization results. The model before adaptation often fails to distinguish between left and right, and even hands and feet. Our method effectively help the model distinguish between different key points on the unlabeled domain.

\begin{table}[htbp]
\caption{PCK on task \textit{SURREAL}→\textit{Human3.6M}. Sld: shoulder, Elb: Elbow. }
\vspace{-5pt}
\addtolength{\tabcolsep}{-4.2pt}
\begin{center}
\begin{tabular}{l|cccccc|l}
\hline
 Method      & Sld & Elb & Wrist & Hip & Knee & Ankle & Avg \\ \hline
 ResNet101 \cite{SimpleBaselines} &     69.4    &  75.4     &    66.4   &  37.9   & 77.3    &   77.7    &  67.3  \\
 DAN \cite{DAN} &     68.1     &  77.5   &  62.3     &  30.4   &   78.4   &   79.4    &   66.0 \\
DANN \cite{DANN} &     66.2     &  73.1   &    61.8   &  35.4   &   75.0   &   73.8    &    64.2\\
MCD \cite{MCD} &      60.3    &  63.6   &  45.0     &   28.7  &  63.7    &  65.4     &  54.5  \\
DD \cite{MDD}    & 71.6   & 83.3    &  75.1   &  42.1  &    76.2       & 76.1  & 70.7  \\
\textbf{RegDA (ours)}       &     \textbf{73.3}     &   \textbf{86.4}    &   \textbf{72.8}   &  \textbf{54.8}   &  \textbf{82.0}    &  \textbf{84.4}     &   \textbf{75.6}  \\
\hline
Oracle      &     95.3     &   91.8    &  86.9     &  95.6   &  94.1    &  93.6  &  92.9    \\ 
\hline
\end{tabular}
\end{center}
\label{result: surreal2human}
\vspace{-10pt}
\end{table}

\begin{table}[htbp]
\caption{PCK on task \textit{SURREAL}→\textit{LSP}.  Sld: shoulder, Elb: Elbow. }
\addtolength{\tabcolsep}{-4.2pt}
\vspace{-5pt}
\begin{center}
\begin{tabular}{l|cccccc|l}
\hline
 Method      & Sld & Elb & Wrist & Hip & Knee & Ankle & Avg \\ \hline
ResNet101 \cite{SimpleBaselines} &         51.5     & 65.0      &  62.9    &  68.0  &  68.7    &   67.4    &  63.9     \\
 DAN \cite{DAN} &      52.2    &  62.9   &  58.9     &  71.0   &  68.1    &  65.1     &   63.0 \\
DANN \cite{DANN} &     50.2     &  62.4   &  58.8     &  67.7   &   66.3   &   65.2    &  61.8  \\
MCD \cite{MCD} &      46.2    &  53.4   &  46.1     & 57.7    &  53.9    &  52.1     &  51.6  \\
DD \cite{MDD}    &     28.4     &  65.9   &  56.8     &  75.0   &  74.3    &   73.9    &   62.4 \\
\textbf{RegDA (ours)}      &      \textbf{62.7}    &   \textbf{76.7}    &  \textbf{71.1}     & \textbf{81.0}    &   \textbf{80.3}   &    \textbf{75.3}    & \textbf{74.6}\\
% \hline
% Oracle      &     81.9     &   73.1    &   66.5    &  81.3   &  74.9    &   70.6    &   74.7  \\
\hline
\end{tabular}
\end{center}
\label{result: surreal2lsp}
\vspace{-10pt}
\end{table}

% 可视化adaptation前后的结果

\section{Conclusion}
In this paper, we propose a novel method for unsupervised domain adaptation in keypoint detection, which utilizes the sparsity of the regression output space to help adversarial training in the high-dimensional space. 
We use a spatial probability distribution to guide the optimization of the adversarial regressor and perform the minimization of two opposite goals to solve the optimization difficulties.
Extensive experiments are conducted on hand keypoint detection and human keypoint detection datasets. Our method is better than the source only model by a large margin and outperforms state-of-the-art DA methods.

{\small
\bibliographystyle{ieee_fullname}
\bibliography{egbib}
}

\end{document}